# Prompting ChatGPT for Translation: A Comparative Analysis of Translation Brief and Persona Prompts

Sui He
School of Culture and Communication
Swansea University
United Kingdom
`sui.he@swansea.ac.uk`

## Abstract

Prompt engineering has shown potential for improving translation quality in LLMs. However, the possibility of using translation concepts in prompt design remains largely underexplored. Against this backdrop, the current paper discusses the effectiveness of incorporating the conceptual tool of "translation brief" and the personas of "translator" and "author" into prompt design for translation tasks in ChatGPT. Findings suggest that, although certain elements are constructive in facilitating human-to-human communication for translation tasks, their effectiveness is limited for improving translation quality in ChatGPT. This accentuates the need for explorative research on how translation theorists and practitioners can develop the current set of conceptual tools rooted in the human-to-human communication paradigm for translation purposes in this emerging workflow involving human-machine interaction, and how translation concepts developed in translation studies can inform the training of GPT models for translation tasks.

## 1 Introduction

Translation quality is a pivotal topic in the field of machine translation. The development of Large Language Models (LLMs) and the popularization of ChatGPT since its public launch in November 2022 have attracted scholarly interests in improving the quality of translation outputs generated by LLMs. Efforts to improve the quality of these translations have involved both fine-tuning and prompt engineering. Despite these efforts, the performance of popular LLMs in executing translation tasks remains suboptimal, particularly when compared with professional translations used in the language service industry (see, for example, Jiao et al., 2023). Therefore, the task for enhancing the performance of LLMs in conducting translation tasks continues to be an on-going effort.

Compared to fine-tuning, prompt engineering provides greater accessibility for ordinary users with translation needs, especially those who operate on the user interfaces of LLMs such as ChatGPT. Most research on prompt engineering for translation purposes draws on concepts such as zero-shot learning rooted in Natural Language Processing (NLP) by feeding sample translations in the context window. In comparison, the possibilities for integrating translation concepts and strategies have received little attention.

From the perspective of advancing translation studies, consolidating the synergy between humans and machines in achieving translation goals at a professional level is crucial. As Lee (2023) rightly notes, "translation as an event can no longer be restricted to translating as an act, given that AI and other communicative modalities will increasingly be drawn into and embedded within the workflow." For the development of translation research, since most translation concepts are anchored in human-to-human communication, it becomes essential to evaluate their efficacy in the emerging workflow with human-machine communication involved, thereby strengthening the disciplinary foundation of translation studies in this novel context. For translation practice, enhancing our understanding of prompt engineering for translation could inform the approach we take for translator training in the changing landscape. As highlighted in a recent work trend report by Microsoft (2023), 82% of leaders from various sectors stated that their employees will need new competencies – such as AI delegation via prompts – to prepare for the expansion of AI.

In the background, this research investigates the effectiveness of incorporating the notion of "transla-tion brief" and the translator/author dichotomy into prompt design, as an attempt to explore the potential of using conceptual tools rooted in translation studies for improving the quality of LLM-generated translations. In this study, ChatGPT is chosen for its popularity among general users and its user-friendly interface that accommodates individuals with limited computing expertise. Based on two sets of experiments, this research seeks to answer two questions specific to the scope of the current study: 1) Compared to a basic translation command, does a prompt containing information included in a typical translation brief help improve the quality of translation outputs? 2) Drawing on the persona feature of ChatGPT, does assigning the role of "translator" make a difference to the translation quality, with the basic instruction and the role of "author" as reference points?

## 2 Literature Review

In the guidance for prompt design in ChatGPT published by OpenAI,[1] six strategies are suggested for creating effective prompts. Even though there is an improvement in the content of this guideline when compared to its earlier version where only three generic strategies (i.e., show and tell, provide quality data, check your settings; accessed April 2023) were listed, OpenAI has not yet published any specialized guidance on prompt design for translation purposes in ChatGPT. Nonetheless, scholarly efforts have been made to address this issue.

In the literature, most of the research focuses on prompting GPT models or other LLMs through APIs (e.g., Vilar et al., 2022; Hendy et al., 2023; Zhang et al., 2023; Zhu et al., 2023). However, a small number of studies have also explored prompt engineering for translation tasks specifically through the user interface of ChatGPT, drawing on different linguistic concepts. Within this niche area, two main threads have emerged: one is centered around specific translation problems and the other features a more holistic approach.

Starting with those targeting specific translation problems, Gu (2023) is the only one in the literature to date. Drawing on the default model (GPT-3.5) of ChatGPT, the author utilizes the "in context learning" capability of ChatGPT (i.e., remembering what has been mentioned in the chat) to "teach" it how to translate attribute clauses. Specifically, a translation strategy commonly adopted by translators to render attributive clauses from Japanese into Chinese was used by the author to design a set of prompts: "What is the noun modified by the attributive clause in the following sentence?", "Place the noun modified by the attributive clause in the subject position of the attributive clause. And then separate [SOURCE SENTENCE] into two sentences", and finally "Translate the following sentence to Chinese: [SEPARATED SOURCE SENTENCE]". Although this prescriptive application of a standalone translation strategy fails to take into consideration the dynamic context of handling attributive clauses, this paper presents a very interesting attempt to bring translation strategies into the horizon of prompt engineering.

Turning to the literature which investigates translation at a contextual level, key concepts tested in this group include "domain", "task", "part of speech", "discourse", and "pivot language" – all of them are well-established topics in translation studies but they have been used in a rather ambiguous way in these works. For instance, Peng et al. (2023) propose the concept of "task-specific prompts" (i.e., "you are a machine translation system") in their experiment, without concrete instructions on what to expect from a so-called "machine translation system". The rationale behind this design, according to the authors, rests in the assumption that ChatGPT has been fine-tuned as a conversation system instead of a machine translation system, and this might have limited the translation ability of ChatGPT. Nonetheless, the effectiveness of altering a fine-tuned chatbot into a machine translation system with a single prompt line in the user interface remains questionable. Additionally, the authors test the efficacy of "domain-specific prompts" (e.g., information about the topic or genre of the ST, such as bio-medical or news-style) by providing ChatGPT with both right and wrong domain information of the ST. This design of using wrong domain information, from the perspective of translation studies, requires careful justification. The results, measured via automated machine translation quality evaluation metrics, suggest that providing task and correct domain information can indeed enhance ChatGPT's translation performance.

Another case in point is Gao et al. (2023). The authors introduce language direction, domain information, and part-of-speech information to their prompt design. Similar to the definition of "domain" in Peng et al. (2023), the authors include information about genre (i.e., news, e-commerce, social, and conversational texts) in their experiments. These prompts were run through five different settings to test their efficacy. The results from automatic metrics further validate the usefulness of domain-related information in prompt engineering for translation tasks. Notably, although the outcome of introducing part-of-speech information in prompts was not promising, it suggests an intention to include grammatical segmentations into prompt design, which echoes the problem-oriented approach to enhancing translation quality, as mentioned above in Gu (2023). An interesting observation made by the authors regarding language direction lies in the disparity between high-resource languages and low-resource languages: domain information appears to enhance machine translation quality for high-resource languages but fails to demonstrate a comparable impact on low-resource languages.

---

[1] https://platform.openai.com/docs/guides/completion/prompt-design

To understand the issue related to high versus low resource languages, Jiao et al. (2023) propose a strategy called "pivot prompting". This notion, bearing similarities to the concept of relay translation, involves instructing ChatGPT to translate the ST into a high-resource language prior to translating it into the target language. Even though the basic prompts were generated by ChatGPT itself without further tweaks, the idea of relay translation turned out to be useful in improving translation quality between distant languages, as the results reported by the authors suggest.

Regarding the topic of context and discourse in translation, whilst all studies mentioned above focus on prompt design for translation at the level of single sentences or small sentence clusters, Wang et al. (2023) take a step forward to the document level. They put forward the concept of "discourse-aware prompts", introducing discourse as an evaluation criterion for assessing the quality of prompts in ChatGPT. To identify the best discourse-aware prompt, the authors evaluate a set of basic prompts generated by ChatGPT with two discourse-oriented metrics: one focuses on terminology consistency and another on the accuracy of zero pronoun translation. As can be seen from the design, discourse here is used in its micro sense as document-level coherence. Macro discoursal information, such as the function of the ST and target audience, is not taken into consideration when designing the prompts.

The most relevant research to date, drawing on a contextualized approach inspired by translation concepts, is reported by Yamada (2023). There are two sets of experiments in this research. First, the author adopts two concepts – purpose of the translation and target readers – for prompting ChatGPT (GPT-4) to translate, simulating a real-life translation commission for ChatGPT. Instead of providing information about the purpose and target readers, the author designed a prompt that asks ChatGPT to find the information itself: "Translate the following Japanese [source text] into English. Please fulfill the following conditions when translating. Purpose of the translation: *You need to fill in*. Target audience: *You need to fill in*. [source text] *You need to fill in*." In the segments shown in italics, the author specifies the information that ChatGPT needs to fill in before generating the translation. Second, the concept of dynamic equivalence is utilized, feeding into ChatGPT as a translation strategy alongside a sample translation of a different source text through in-context learning. This combined approach complicates the task of determining whether the concept of "dynamic equivalence" or its illustrative examples play a more significant role in the efficacy of the prompt. To assess the overall effectiveness of this prompt, the author uses cosine similarity of vectors as indicators for semantic proximity and a detailed qualitative evaluation conducted by the author himself, with reference translations generated by DeepL, Google Translate, and ChatGPT (with default prompt "Translate to English"). The author reports that "incorporating the purpose and target readers into prompts indeed altered the generated translations" and that "this transformation […] generally improved the translation quality by industry standards". This research features a very interesting attempt to "teach" ChatGPT to "think" and "act" like a translator via prompts, revealing the potential for training ChatGPT with knowledge generated by translation scholars.

Overall, the current landscape of prompt design in ChatGPT features important attempts to enhance its capability in executing translation tasks. However, a critical issue with these endeavors lies in the fact that the concepts being used in the prompts (e.g., "news-style") are too general to be informative, and some of the approaches (e.g., the out-of-context application of prescriptive translation strategies) bear striking resemblances to what happened in the early days in translation studies. The design of prompts shows that these research efforts have touched upon some key conceptual tools for translation, revealing the potential benefit that translation concepts can bring for enhancing LLMs' performance in generating professional level translations.

## 3   Research Design

Building on the effectiveness of introducing contextual and domain-specific information as demonstrated in the literature, this paper investigates prompt design in light of two conceptual tools rooted in translation research: first, "translation brief" as featured in the functionalist approach to translation; second, the "author-translator" dynamic given the persona-matching feature of ChatGPT.

## 3.1 Prompt design

In total, four prompts were tested in this pilot study, including one basic prompt functioning as a baseline for comparison, and three other prompts featuring three keywords in the scholarship of translation studies: translation brief, author, and translator.

For the basic prompt, because the aim is to evaluate the translation performance of ChatGPT in a professional setting, information included is: 1) a translation command, 2) the target language, and 3) the purpose for professional use, as one would set out in a translation commission. This information was also included in the three other prompts.

For the translation brief prompt, factors including intended text functions, addressees, time and place of text reception, the medium, and the motive (Munday et al., 2022) were included.

For the author-translator dynamic embedded in the source-target dichotomy, discussions on these two roles and their implications for translation studies have been well documented in the trajectory of translation research. Assigning a persona to ChatGPT is a key feature of the GPT models, and this provides the possibility of incorporating this pair of keywords into prompt design.

Furthermore, the temperature is set at 0.5 for each prompt to constrain the degree of creativity that ChatGPT can potentially exhibit, mimicking the freedom that translators can potentially take in translating articles of this genre in real-life scenarios.

An overview of the four prompts is presented in Table 1 below.

| Prompts | Content |
| --- | --- |
| Basic | Please translate the following text from English into Chinese Mandarin. The translation is intended for professional use. Top_p=0.5 |
| TransBrief | Please translate the following text from English to Chinese Mandarin. The paragraph is taken from a popular scientific article published in *Discover Magazine*. The translated version will be published on the *Scientific American* website in 2023 for professional use. The author of the original text is a well-known science writer, and the target audience for the translation consists of educated individuals interested in popular science. The original text aims to communicate recent research in mathematics that explores the fundamental principles of time travel. Top_p=0.5 |
| Author | You are a professional popular science author. Please translate the following text from English into Chinese Mandarin. The translation is intended for professional use. Top_p=0.5 |
| Translator | You are a professional popular science translator. Please translate the following text from English into Chinese Mandarin. The translation is intended for professional use. Top_p=0.5 |

**Table 1**. Prompt Overview.

## 3.2 Text generation

The source text (ST) selected for the study is a popular scientific article published on the website of the *Discover* magazine in December 2021.[2] This genre is chosen for its dual emphasis on maintaining rigorous scientific accuracy and employing a nuanced narrative style, which requires authors and translators to communicate scientific knowledge in a manner that is both accessible and engaging to their respective audiences. The article, titled "A Major Time Travel Perk May Be Technically Impossible", was written by Cody Cottier, a professional popular science writer. Drawing on a publication of researchers based at the University of Queensland in Australia, the popular scientific article provides accessible and engaging information about time travel for an English-speaking audience interested in but not necessarily have specialized knowledge of this topic.

The selection criteria for the ST are influenced by multiple factors: first, the May 2023 version of ChatGPT utilized in this research has a knowledge cut-off date of September 2021; second, its token capacity (i.e., how many texts it can handle in a single input) is limited; third, the ST should be a professional text; and fourth, a published translation which can serve as a reference document for automatic quality evaluation should be available. To satisfy these basic requirements, the ST is manually checked against the lexical updates on the *Oxford English Dictionary* website[3] to ensure it does not contain any neologisms coined after September 2021. Also, the length of the ST (1253 words) is manageable for ChatGPT. The authoritative status of *Discovery* in popular science journalism and the availability of a published Chinese translation by *Huanqiukexue* – a renowned popular science magazine in China – further make the ST a suitable choice.

---

[2] https://www.discovermagazine.com/the-sciences/a-major-time-travel-perk-may-be-technically-impossible
[3] https://www.oed.com/information/updates

The model used in the experiment is GPT-4, accessed via the user interface of ChatGPT. Compared to GPT-3.5, this model has demonstrated superior performance in machine translation (Jiao et al. 2023; Wang et al. 2023). All translation outputs were generated by the 24 May 2023 version of ChatGPT. Markdown language was used in the ST to help ChatGPT differentiate headings from main texts and infer the structure of the ST based on the text formatting. Delimiters were used to define the beginning and the end of the ST. Since ChatGPT cannot generate a complete translation in a single response, the prompt "go on" was used to resume the translation command. To assess the consistency of translation outputs generated by the prompts, each prompt was tested three times using a sample sentence from the ST. The outputs were then manually examined by the author for consistency, with a rating scale ranging from 0 to 3, where 0 denotes "Professionally Unusable", 1 denotes "Professionally Usable with Major Modification", 2 denotes "Professionally Usable with Minor Modification" and 3 denotes "Professionally Usable". All four prompts consistently produced similar translations based on the rating. The fourth output from each prompt was selected as the sample for the analysis.

The translation published in *Huanqiukexue* was labeled as TT1, and four machine translations were labeled as TT2 (Basic), TT3 (TransBrief), TT4 (Author) and TT5 (Translator), where TT stands for Target Text. The summary of the word count of Chinese characters in each TT (mean ≈ 2430, standard deviation ≈ 88) is presented in Table 2 below, offering an idea about the size of the translations.

| TT1 | TT2 | TT3 | TT4 | TT5 |
|---|---|---|---|---|
| 2602 | 2379 | 2374 | 2369 | 2424 |

**Table 2**. Word Count of the TTs in Chinese characters.

### 3.3 Quality evaluation

Both automatic and human evaluations were conducted to assess the quality of the translation outputs. Two quality evaluation metrics were adopted in this study: BLEU (Papineni et al., 2002) and COMET-22 (Rei et al., 2022). COMET-22 was chosen for its outstanding performance in WMT22 Metrics Shared Task and availability (Freitag et al. 2022). Although BLEU has been criticized heavily for its reliability, it has been chosen as a reference to triangulate results generated by COMET-22 and human evaluations.

To prepare the ST and TTs for automatic evaluation, SDL Trados Studio 2022 was used to align the source and target segments. In total, 66 aligned segments were generated for each ST−TT pair. These aligned texts were then converted into plain text files for BLEU and compiled in an Excel workbook for COMET-22. For BLEU, the text files were processed through the user interface developed by Tilde.[4] For COMET-22 (`wmt-comet-da`[5]), the metric was run in Python to generate results.

Human evaluations were conducted for qualitative analysis. Four evaluators contributed; all of them are university lecturers based in the UK, who have extensive theoretical and practical knowledge of English−Chinese translation. The evaluators were invited to grade all five TTs (four machine translation outputs and one human translation), without knowledge of which ones were machine-generated translations. Ethical approval was granted by the Humanities and Social Sciences Ethics Committee of Swansea University, before the collection of evaluations (research ethics approval number: 2 2023 6610 5739). Each evaluator was provided with an information sheet and a consent form before taking part in the evaluation.

The grading form designed for human evaluation is different from the metrics typically used in the development of machine translation systems, such as those outlined by Freitag et al. (2022). Instead, it was designed from a translation studies perspective to encourage evaluators to assess the translations on a textual level, following a "top-down approach" (Han 2020) to obtain a relative ranking of the TTs. Furthermore, to capture individualized responses regarding the strengths and weaknesses of translations, fixed rubrics containing guided scales were intentionally omitted. This decision stems from the understanding that translation is more than technical transfer of information and that evaluators are not only experienced translation assessors but also readers within this context. Traditional evaluation scales often focus on aspects such as "accuracy" and "adequacy" to ensure replicability and other concerns in machine translation quality assessments. However, such criteria can oversimplify the nuanced nature of translation as a social activity. Discussions on good versus bad translations are not the primary concern in translation studies; rather, since the cultural turn in the 1990s, translation has been discussed as a socio-historical phenomenon. This viewpoint allows individual interpretations of a ST to be manifested through the medium of translation, which can influence social narratives in another language or culture. This is also true for popular scientific articles embedded with tactical narratives. Traditional criteria reduce the complex social dynamics of translation to mere encoding and decoding of static information, which does not reflect how audiences engage with translated works in real-life scenarios. Without an evaluation scale that comprehensively considers reader reception, the method adopted in this study allows evaluators the freedom to express their opinions without interference. This approach provides a more accurate reflection of the real-world reception of translations. Admittedly, this might not be the case for some domains, and it would be beneficial to have a reader-oriented scale to use, especially at this point of AI development, but it is beyond the scope of the current project.

---

[4] https://www.letsmt.eu/Bleu.aspx
[5] https://huggingface.co/Unbabel/wmt22-comet-da

Based on semantic and structural information embedded in the ST, it was divided into ten segments to create a reading flow for evaluators that resembles the natural reading habits of humans, rather than soliciting evaluations for the sake of evaluation. The source and target segments were aligned in ten blocks in the grading form for easier comparison. Numerical grading boxes (based on a scale of one to ten, with one being the worst and ten the best) and optional free text boxes were provided for each segment. An overall rating block was also included at the end of the grading form.

Figure 1. Human Evaluation Grading Form – An Example.

In total, each evaluator recorded eleven grades for each TT. For segment grades, the averages were taken for each segment in order to obtain the relative ranking, detailed information can be found in section 4.2.

## 4 Findings and Discussion

Results from the automatic evaluation metrics and human grading forms provide complementary insights into the quality of the generated TTs, indicating the efficacy of each prompt. This section starts with the results of the two automatic metrics, before delving into human evaluation results.

### 4.1 Machine evaluation

BLEU and COMET-22 provide scores at both segment and whole text levels. Therefore, each TT yields 67 data points (66 segment scores and one overall score). Table 3 presents the overall scores for the four AI-generated TTs in BLEU and COMET-22, with the rankings shown as superscripts.

| Metric | TT2 | TT3 | TT4 | TT5 |
|---|---|---|---|---|
| BLEU | 4.03[2] | 3.31[4] | 3.9[3] | 7.89[1] |
| COMET | 0.8233[3] | 0.8224[4] | 0.8247[2] | 0.8296[1] |

Table 3. BLEU and COMET Overall Scores and Rankings.

In both metrics, TT5 (translator) achieved better performance than the three other TTs, and TT3 (translation brief) was ranked the lowest quality. The rankings of the four TTs in BLEU and COMET, however, are different with regard to TT2 and TT4, as shown in Table 3. In general, TT5 (translator) achieved the highest rank across the two metrics, with TT2 (basic) and TT4 (author) following behind. TT3 (translation brief), however, hit the less optimal ground.

Additionally, the differences of the segment scores were tested between TT2 (basic) and TT3 (translation brief), TT2 (basic) and TT5 (translator), and TT4 (author) and TT5 (translator). Wilcoxon matched-pairs signed-ranks tests were employed due to the non-normal distribution of data. Statistical analyses were conducted in Python using the pandas (McKinney, 2010) and scipy.stats (Virtanen et al., 2020) packages.

Results show that none of the differences are statistically significant. In BLEU, for the translation brief prompt, the overall score for TT2 (4.03) is higher than TT3 (3.31) by approximately 21.75%. However, the difference, based on the segment scores, is not statistically significant (p = 0.126, effect size = 1.21). For the persona group, the overall score for TT5 (7.89) is higher than TT4 (3.9) by approximately 102.3%. Yet, the difference at a segment level is also not statistically significant (p = 0.785, effect size = 4.06). For the COMET-22 segment scores, results are also insignificant: for TT2 and TT3, the p-value is 0.7853 (effect size = 13.30) and for TT2 and TT5, the p-value is 0.190 (effect size = 0.618). For TT4 and TT5, the p-value is 0.2501 (effect size = 12.73).

These statistically insignificant results could be attributed to the fact that both BLEU and COMET-22 were not initially designed to evaluate the effectiveness of individual prompts within a system. Another potential explanation is that the published translation may not be a suitable reference document for these automatic metrics: even though the omissions and relocations of information in the published translation could potentially enhance its overall communicative effect, this type of translation behavior does not align with the algorithms embedded in BLEU or COMET-22. Equally, it could also be the case that the information typically provided in translation briefs does not assist ChatGPT in producing better translations in the same way that it assists human translators. To have a better insight into these issues, the following section reports on human evaluation results.

## 4.2 Human evaluation

At a document level, the overall grades given by the evaluators and the standard deviations are listed in Table 4 below. No statistical tests were conducted to assess the significance of differences due to the small number of data points generated in this set of evaluations.

| Reviewer<br>TT No. | 1 | 2 | 3 | 4 | Avg | Rank |
|---|---|---|---|---|---|---|
| **TT1** | 7 | 9 | 9 | 5 | 7.5 | 1 |
| **TT2** | 5 | 4 | 4 | 5 | 4.5 | 4 |
| **TT3** | 4 | 4 | 4 | 6 | 4.5 | 4 |
| **TT4** | 4 | 6 | 6 | 6 | 5.5 | 3 |
| **TT5** | 5 | 6 | 6 | 6 | 5.75 | 2 |

**Table 4**. Human Evaluation: Overall Scores with Rankings.

TT1, the published version, received the highest ranking on average. Interestingly, among the four machine translations, human evaluation results also show a preference for TT5 (translator) over the three other prompts. The rankings of TT4 and TT5 also indicate that assigning a persona to ChatGPT tends to enable it to produce a better translation, compared to the translations produced with the basic and the translation brief prompts.

Whilst the overall grades of TT2 and TT3 are identical, the average grades of individual segments reveal a difference between the two. At the segment level, the ten segments add up to a total score of 100. Given that the evaluators for the TTs are the same, taking the average of the segment scores helps to cancel out the individual preferences of each evaluator as a result of maintaining the relative ranking of each translation, based on the assumption that all evaluators are consistent within their own scoring schemes.

Table 5 shows the sums and averages of segment scores for each TTs below. As can be seen in Table 5, the performance of TT5 is the highest among the four prompted outputs, followed by TT2, TT4, and TT3, and these data are in line with the overall scores for the TTs in the automatic metrics.

| Reviewer<br>TT No. | 1 | 2 | 3 | 4 | Avg | Rank |
|---|---|---|---|---|---|---|
| **TT1** | 66 | 73 | 89 | 53 | 70.25 | 1 |
| **TT2** | 47 | 54 | 52 | 57 | 52.5 | 3 |
| **TT3** | 41 | 48 | 58 | 59 | 51.5 | 5 |
| **TT4** | 38 | 56 | 57 | 58 | 52.25 | 4 |
| **TT5** | 46 | 54 | 57 | 60 | 54.25 | 2 |

**Table 5**. Human Evaluation: Accumulated Sums of Segment Scores with Rankings.

Moving on to the comments given by the evaluators for the TTs, for the machine translation outputs, three keywords emerged among the issues pointed out by the evaluators: fluency/naturalness, reader-friendliness, and accuracy.

First, comments on the issues of fluency and naturalness suggest problems associated with syntax, collocation, and lack of creativity in rendering expressions that are not commonly seen in Chinese languages. For instance, the verb "lead" in segment [2] "the past will likely always *lead* to the same future" was translated as 导致 (lead to a result), 导向 (lead to a direction, usually as a noun) and 引导到 (to guide to) by ChatGPT, which were commented by evaluators on lexical choices that "tend to be made at a surface level".

Second, taking reader experience into consideration, comments were made on the literal translations of source segments by ChatGPT as "may distract or discourage the readers", "I'm not sure what this is supposed to mean", "difficult to follow", and "this [translation segment] is not clear". The semantic emphasis of Chinese, especially the use of particles to indicate tenses, also tends to be ignored in the machine translations, as an evaluator mentioned.

Third, two inaccurate translations have been identified by evaluators. For instance, there is one omission example identified by evaluators: a piece of information included in brackets in the ST was omitted in TT2, which led to a fluency issue as an evaluator pointed out, quoting "The text reads more fluently when this clause is included as an organic part of the sentence." Another case in point is related to terminology accuracy in context. Segment 5 in the ST starts with "no one knows whether time travel is physically possible", and "physically" here was rendered as 物理上 (literally, regarding Physics) in all four ChatGPT translations. As an evaluator notes, this translation "makes sense but is not as accurate and easy to understand as 技术上" (literally, technically), as seen in the human translation.

For the human translation, on the other hand, most comments are related to the issue of accuracy, specifically with regards to the deviation of meaning and omission cases. This issue, as shown in the comments, is mainly related to the creative modifications of the original text made by the human translator. Creativity, in this case, presents itself as a double-edged sword. For instance, the creative translation of the title was highlighted by evaluators, both as strengths and weaknesses from different perspectives. For one evaluator, the human translation of the title was favored by one evaluator, quoting "I think 'major time travel perk' is difficult to render in Chinese […] Strictly speaking, TT1 did not follow the ST but adopted a more creative solution. I really like this translation. This sounds exactly like the title of an article you'd read in a popular science magazine." Notably, another evaluator also commented on the positive impact that the freedom shown by the human translators in rendering the title, but at the same time, the negative impact was also pointed out: "It is in the style of title to start with; it conforms less closely to the wording of the ST but incorporates an understanding of the whole article. This is something an experienced translator with good Chinese skills would do or would aim for, at least. Nevertheless, this translation apparently suggests the main purpose of the article is to introduce the physics of time-travel, which is slightly off target." Similarly, in another segment, the translation of a subheading "Time Without Beginning" as 没有起点的故事 (literally, a story without beginning), was pointed out by one evaluator as inaccurate, due to the mistranslation of "time" as "story".

### 4.3 Summary of Findings

Overall, based on automatic evaluation metrics and human evaluation scores, the rankings of the TTs show that the basic prompt led to better performance of ChatGPT in translation than the prompt including information typical of a translation brief. For the employment of personas to guide ChatGPT, assigning the role of a translator is more effective than the basic prompt and assigning the role of an author, and it has actually led to the best performance among the four prompts tested. For human evaluation comments, it is shown that while the main issues with ChatGPT-generated translations rest on the issues of fluency and naturalness, the comments for the published translation focus mainly on accuracy, mostly resulting from the creativity and stylistic choices shown in the text.

These findings suggest that providing the information contained in a typical translation brief used in human-to-human communication for translation commissions does not necessarily lead to a better performance of ChatGPT in completing translation commands, and that assigning ChatGPT with the role of a translator appears to have a better result than assigning the role of an author or just using a basic prompt.

## 5 Conclusion

This research explores the efficacy of integrating concepts developed in translation studies into prompting ChatGPT for translation tasks. By evaluating the outputs generated by ChatGPT under four different prompts, it seeks to provide insights into the effectiveness of giving a translation brief to ChatGPT and assigning ChatGPT the personas of an author and a translator. Findings show that assigning the persona as a translator allowed ChatGPT to achieve the best performance among the four prompts, and that the translation generated by ChatGPT using the translation brief prompt received the lowest ranking. This indicates that the classical settings of a translation brief, aiming at human-only workflow, might not work as well as one would expect in a human-machine workflow. However, it would be necessary to revisit the conceptual tools developed in translation studies, considering the development of translation technology and the changing landscape in the industry, so as to further consolidate the relevancy and credibility of translation studies as a discipline. Similarly, training GPT models using aligned source and target texts, paired with translation briefs, and exploring other concepts developed in translation studies could be potentially beneficial.

There are some limitations of the current research. For instance, when testing the consistency of the prompts based on the translation outputs generated by ChatGPT, involving multiple raters, and conducting an inter-rater reliability test would be helpful. Additionally, a reader-centered human evaluation metrics and interviews with human evaluators would have been a good complement to the information based solely on the textual analysis of evaluators' comments extracted from the grading form. In addition, using document-level quality evaluation metrics might also strengthen the discussion of the results.

As mentioned in the introduction, this research only provides partial insights into the two general research questions, based on the data collected in this experiment. To further develop this line of research, different prompts conveying information about translation concepts could be examined, across various genres, assessed with a human evaluation scale closer to the reality of translation reading by a larger number of human evaluators. This approach would generate more data, allowing for replication and statistical testing to enhance reliability. Additionally, with the development of Generative AI, research into other LLMs for translation purposes could offer valuable comparative insights for both practitioners and researchers in the field.

Thinking forward, as Hendy et al. (2023) rightly note, although GPT models have promising potential in machine translation, their performance remains underexplored compared to commercial machine translation systems. LLMs are developing rapidly as we write. By extending the scope of translation studies from human-to-human communication to human-machine communication, translation researchers can help to co-shape the future of machine translation and theorize the practice of translation in the new era.

**Acknowledgements:** I wish to express my gratitude to Dr. Caiwen Wang (UCL/University of Westminster), Dr. Min-Hsiu Liao (Heriot-Watt University), Dr. Yu-Kit Cheung (University of Manchester), and Dr. Yunhan Hu (Durham University) for their valuable evaluation of the translations. Additionally, I extend my thanks to Professor Tong King Lee and the anonymous reviewers for providing insightful feedback on the earlier versions of this manuscript.